\documentclass[10pt,twocolumn,letterpaper]{article}

\usepackage{wacv}              

\usepackage{graphicx}
\usepackage{amsmath}
\usepackage{amssymb}
\usepackage{booktabs}
\usepackage{algorithm}
\usepackage{algpseudocode}

\usepackage[pagebackref,breaklinks,colorlinks]{hyperref}

\usepackage[capitalize]{cleveref}
\crefname{section}{Sec.}{Secs.}
\Crefname{section}{Section}{Sections}
\Crefname{table}{Table}{Tables}
\crefname{table}{Tab.}{Tabs.}

\def\wacvPaperID{1600} 
\def\confName{WACV}
\def\confYear{2025}

\begin{document}

\title{MaskVD: Region Masking for Efficient Video Object Detection}

\author{Sreetama Sarkar$^{1}$ \ \ \ \
Gourav Datta$^{1}$ \ \ \ \
Souvik Kundu$^{2}$ \ \ \ \ 
Kai Zheng$^{1}$ \\
Chirayata Bhattacharyya$^{3}$ \ \ \ \
Peter A. Beerel$^{1}$\\
$^{1}$Universiy of Southern California, Los Angeles, USA \ \ \ \ 
$^{2}$Intel Labs, San Diego, USA \\
$^{3}$Indian Institute of Science, Bangalore, India \\
{\tt\small {\{sreetama,gdatta,kzheng44,pabeerel\}@usc.edu} \ \  \tt\small {souvikk.kundu}@intel.com} \ \ \tt\small {chirayatab}@iisc.ac.in}

\maketitle

\begin{abstract}
Video tasks are compute-heavy and thus pose a challenge when deploying in real-time applications, particularly for tasks that require state-of-the-art Vision Transformers (ViTs). Several research efforts
have tried to address this challenge by leveraging the fact that large portions of the video undergo very little change across frames leading to redundant computations in frame-based video processing. In particular, some works leverage pixel or semantic differences across frames, however, this yields limited latency benefits with significantly increased memory overhead. This paper, in contrast, presents a strategy for masking regions in video frames that leverages the semantic information in images and the temporal correlation between frames to significantly reduce FLOPs and latency with little to no penalty in performance over baseline models. In particular, we demonstrate that by leveraging extracted features from previous frames, ViT backbones directly benefit from region masking, skipping up to 80\% of input regions, improving FLOPs and latency by 3.14$\times$ and 1.5$\times$. We improve memory and latency over the state-of-the-art (SOTA) by 2.3$\times$ and 1.14$\times$,  while maintaining similar detection performance.
Additionally, our approach demonstrates promising results on convolutional neural networks (CNNs) and provides latency improvements over the SOTA up to 1.3$\times$ using specialized computational kernels.
\end{abstract}

\section{Introduction}
\label{sec:intro}
With the emergence of continuous mobile vision and the widespread adoption of cameras on wearable and mobile devices, the demand for real-time video processing applications is rapidly increasing. These applications play a crucial role in a wide range of fields, including surveillance, webcams, and autonomous driving \cite{nuscenes2019, Geiger2012CVPR}. The video streams captured by the image sensor are processed by computer vision (CV) models to carry out image understanding tasks such as classification \cite{russakovsky2015ImageNet, cifar}, segmentation \cite{chen2017deeplab, Ronneberger2015UNetCN}, and detection \cite{lin2014microsoftcoco, Geiger2012CVPR}. In particular, Vision Transformers (ViTs) \cite{dosovitskiy2020image} have proven to be highly effective for these tasks. However, their high computational requirements have introduced significant challenges for resource-constrained latency-critical applications, especially in handling high-resolution images sampled at high frame rates (FPS). In particular, there are concerns about energy consumption and bandwidth requirements for transmitting these image frames from the sensors to the CV backend. 

\begin{figure}
    \centering
    \includegraphics[width=0.75\linewidth]{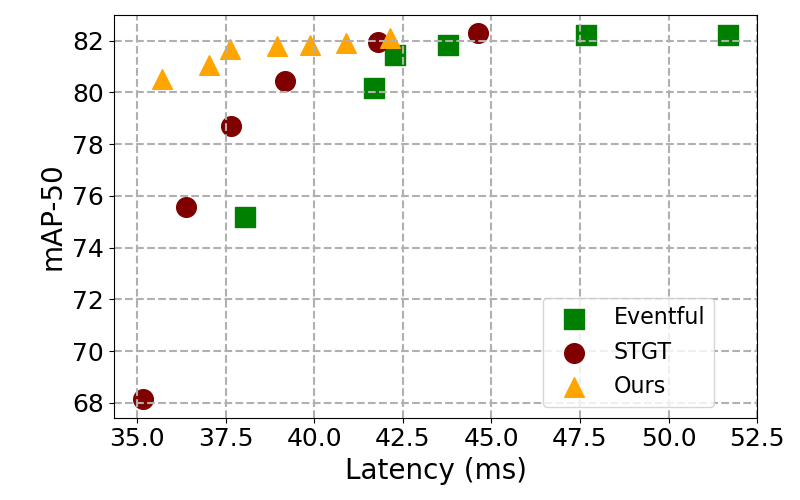}
    \caption{mAP-50 vs. latency comparison of MaskVD with STGT \cite{liSpatiotemporalGated2021} and Eventful-Transformer \cite{Dutson_ICCV_2023}, two SOTA token-dropping methods on ViTs for frame-based video object detection on ImageNet-VID dataset. MaskVD achieves the best mAP-50 latency trade-off.}
    \label{fig:intro}
\end{figure}

Several efforts have been made to reduce computational cost of ViTs, including using simplified attention calculation \cite{wang2020linformer, zhang2023sal}, designing latency-efficient architectures \cite{chen2022mobile, li2022efficientformer} as well as pruning input activations \cite{liang2022evit, bolya2022tokenmerge}. Temporal redundancy in video frames provides an orthogonal direction and potential for significantly higher computation reduction.
In particular, a large portion of videos undergo very little change between successive frames, resulting in redundant computations when frames are processed individually. Previous works exploiting temporal redundancy in videos primarily use delta-based approaches \cite{parger2022deltacnn, Dutson_ICCV_2023}, where pixels or regions are processed only when their difference from the previous frame exceeds a certain delta threshold. These approaches are generally effective for stationary cameras but require additional effort for frame alignment \cite{parger2023motiondeltacnn} when the inputs are captured using moving cameras. Furthermore, they significantly increase the memory footprint since several intermediate feature maps have to be stored in order to compute the associated deltas. In fact, they fail to effectively reduce latency in ViTs \cite{Dutson_ICCV_2023} due to the overhead of additional operations for region selection in intermediate layers. 

We introduce \textit{MaskVD}, a method designed for frame-based video object detection that performs well with both static and moving cameras and can be seamlessly integrated during inference without requiring re-training. In particular, we develop a novel region masking strategy and adapt detection backbones for efficiently reusing features extracted from previous frames. This significantly reduces latency and FLOPs while maintaining similar performance compared to existing methods, as shown in Figure \ref{fig:intro}.

\noindent\textbf{Our Contributions:}
We propose a \emph{region masking strategy} that considers the semantic information in images and the temporal correlation between frames, processing only regions of importance and improving detection performance while reducing computational costs. This input masking strategy can be applied to both CNNs and ViTs alike. The patch selection based on the input mask is performed only at the input of the detection pipeline, unlike existing methods that require re-selection in the intermediate layers. Moreover, our mask can be determined before processing the current frame, thereby saving pixel-readout as well as data transmission energy for the masked regions. This can significantly reduce the bandwidth and total system energy for recently developed in-sensor \cite{in-sensor1,in-sensor2} and near-sensor \cite{pinkhan2021jetcas} computing systems.
We achieve masking up to 80\% of overall input regions with close to baseline detection performance.

In our detection pipeline, we leverage extracted features from previous frames for transformer backbones, storing a single tensor for the windowed transformer blocks, significantly reducing memory over existing methods without degrading performance. We demonstrate results for video object detection using the ViTDet \cite{Li2022vitdet} on two different datasets: ImageNet-VID \cite{russakovsky2015ImageNet}, captured using static camera, and KITTI \cite{Geiger2012CVPR}, captured using moving camera. Our approach performs well both for static and moving cameras and can skip up to $\sim$80\% of the input regions, reducing FLOPs and latency by 3.14$\times$ and 1.5$\times$ with a performance degradation of 0.9\%. We improve latency and memory over the SOTA by 1.14$\times$ and 2.3$\times$, while achieving similar detection performance. We observe higher latency benefits for non-windowed backbones even at lower sparsity, although this comes at a small cost in accuracy.  


We also showcase improved detection performance of up to 1.4\% with CNN backbone using our masking strategy. Since input sparsity does not directly yield computational benefits on GPUs for CNNs, we leverage the specialized computational kernel introduced in DeltaCNN \cite{parger2022deltacnn}. Our region masking approach improves latency by 1.3$\times$ over regular DeltaCNN without input masking. 

\section{Related Work}
\label{sec:related_work}
\noindent\textbf{Efficient Video Inference:}
Several works \cite{parger2022deltacnn, cavigelliCBinferChangebased2017, habibianSkipconvolutionsEfficient2021, xu2018deepcache} have focused on leveraging temporal redundancy between frames for accelerating video inference in CNNs. While DeepCache \cite{xu2018deepcache} skips computations at block level using traditional block matching algorithms, most of the other methods such as Skip-Convolution \cite{habibianSkipconvolutionsEfficient2021}, CBIInfer \cite{cavigelliCBinferChangebased2017}, DeltaCNN \cite{parger2022deltacnn} skip computations at pixel level by creating a pixel mask based on thresholding pixel differences between consecutive frames. However, activation sparsity in CNNs cannot yield actual speedup on general purpose hardware like GPUs. DeltaCNN \cite{parger2022deltacnn} designs specialized CUDA kernels showing latency benefits on GPUs. Unlike CBIInfer \cite{cavigelliCBinferChangebased2017} and Skip-Convolution \cite{habibianSkipconvolutionsEfficient2021} which require caching the input and output feature maps for every convolution layer, Delta-CNN performs delta computations only at the input, and propagates sparse updates through all layers providing actual speedup. These delta-based approaches filter out background regions for static cameras and facilitate processing only objects of interest. However, moving camera sequence need frame alignment prior to delta computation, as demonstrated by MotionDeltaCNN \cite{parger2023motiondeltacnn}.  Our input region masking approach can provide further latency improvements over DeltaCNN by reducing delta computation in large portions of the frame. Additionally, there has been prior research on video inference using compressed data (such as H.264 and HEVC), which demonstrates improved performance due to its higher information density \cite{compressed_video}.  Intuitively, redundant frames can also be skipped based on motion estimation from optical flow \cite{optical_flow_motion} or other motion vector networks \cite{optical_flow_network}; however, that would be computationally expensive. More recently, researchers have exploited temporal redundancy in ViTs \cite{Dutson_ICCV_2023, liSpatiotemporalGated2021}, as detailed in the next subsection.

\vspace{3mm}
\noindent\textbf{Efficient Vision Transformers:}
Efficient ViTs is an active area of research and can be broadly partitioned into two categories: model optimizations and pruning redundant input regions for computational complexity reduction. In the first category, researchers have explored different self-attention alternatives, including Linformer \cite{wang2020linformer}, SAL-ViT \cite{zhang2023sal}, and Performer \cite{choromanski2020rethinking}, that targeted reducing its quadratic compute complexity. Other works presented resource and latency-efficient architectures, including Mobile-former \cite{chen2022mobile} and Efficientformer \cite{li2022efficientformer}. The second category exploits spatial redundancy in non-overlapping input patches, also called tokens \cite{liang2022evit, sarkarECV24, fayyaz2022ats} or input windows \cite{chen2023sparsevit}, consisting of several tokens. Token-level approaches include EViT \cite{liang2022evit}, Adaptive Token Sampling \cite{fayyaz2022ats}, which drop tokens based on self-attention scores, and Token Merging \cite{bolya2022tokenmerge}, which merges tokens based on token similarity. These methods progressively drop or merge tokens along the depth of the network, resulting in a reduced number of tokens feeding into the classification head. This creates a challenge for detection tasks, as the backbone is typically followed by a pyramid network that necessitates the preservation of input dimensions requiring interpolation. 
SparseViT \cite{chen2023sparsevit} performs window-level pruning based on the L2 score of windows with layerwise sparsity ratios determined using evolutionary search. Temporal redundancy in video frames provides a significant opportunity for further computation reduction. Recently, Spatial-Temporal Token Selection (STTS) \cite{wangEfficientVideo2022}, Eventful-Transformer \cite{Dutson_ICCV_2023} and Spatio-Temporal Gated Transformer (STGT) \cite{liSpatiotemporalGated2021} have exploited the spatio-temporal redundancy in video images for efficient video inference in ViTs. While STTS \cite{wangEfficientVideo2022} accelerates video transformers taking entire video sequence as input, STGT and Eventful-Transformer handle individual video frames. However, STGT accelerates linear layers only and uses lossy gating logic leading to accuracy degradation. Eventful-Transformer \cite{Dutson_ICCV_2023}, while demonstrating high reduction in FLOPs, does not yield significant latency benefits and exhibits increased memory overhead. 
In particular, Eventful-Transformer introduces \textit{token gates} and \textit{token buffers} at the input and output of each layer within a transformer block. The \textit{token gate} is responsible for gathering tokens with higher delta differences, while the \textit{token buffer} updates the processed tokens in the corresponding locations of the most recent known reference tensor. Both \textit{token gates} and \textit{token buffers} store a reference tensor from the previous frame, significantly increasing the required memory. Moreover, repeatedly switching tokens between sparse and dense configurations using scatter-gather incurs significant latency overheads \cite{Dutson_ICCV_2023}.  

Our approach requires storing only one feature map for every windowed transformer block and the output feature map of the backbone, thereby reducing memory demand significantly. We further reduce the number of scatter-gather operations, thereby reducing latency.

\section{Approach}
\label{sec:approach}
\subsection{Preliminaries}

ViT models partition the input image into $N$ non-overlapping patches, called \emph{tokens}, and embed each token into an embedding vector of length $L$. A trainable positional embedding is added to each of the token embeddings, which are then passed through a series of transformer encoder blocks, each consisting of a multi-head self-attention (MSA) layer followed by a feed-forward network (FFN). The tokens are mapped into Query ($\mathbf{Q}$), Key ($\mathbf{K}$), and Value ($\mathbf{V}$) matrices, each having a dimension of $\left[N+1, d\right]$, where $d=L/H$ and $H$ is the number of heads in the MSA. Each head performs the self-attention operation shown in Equation \ref{eq:attn}. The FFN is usually a 2-layer network with GELU activation. 
\begin{equation}
    {\mathbf{Attention}}(\mathbf{Q}, \mathbf{K}, \mathbf{V}) = {\rm Softmax}(\frac{\mathbf{Q} \mathbf{K}^T}{\sqrt{d}}) \mathbf{V}.
    \label{eq:attn}
\end{equation}

ViTDet~\cite{Li2022vitdet} demonstrates SOTA performance on detection tasks leveraging plain transformer backbones. The backbone in VitDet \cite{Li2022vitdet} consists of windowed attention blocks (W-MSA), which divide the tokens into non-overlapping windows. The self-attention computation, given by Equation \ref{eq:attn}, is constrained within local windows in W-MSA blocks, lowering the computational cost of self-attention. W-MSA blocks are interleaved with a few global self-attention (MSA) blocks to facilitate information propagation between windows. 

\subsection{Constructing Input Mask}
\begin{figure}[htbp]
  \centering
  \begin{subfigure}{0.3\linewidth}
    \centering
    \includegraphics[width=0.9\linewidth]{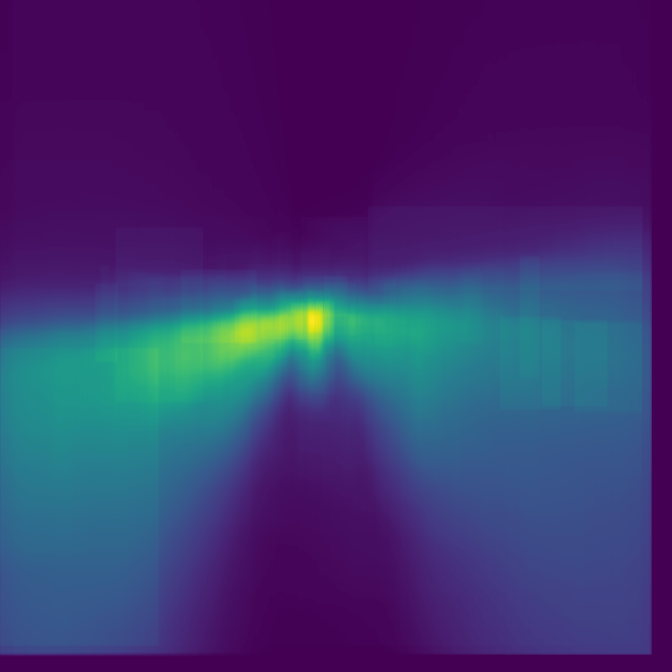}
    \caption{}
    \label{fig:heatmap_kitti}
  \end{subfigure}
  \begin{subfigure}{0.3\linewidth}
    \centering
      \includegraphics[width=0.9\linewidth]{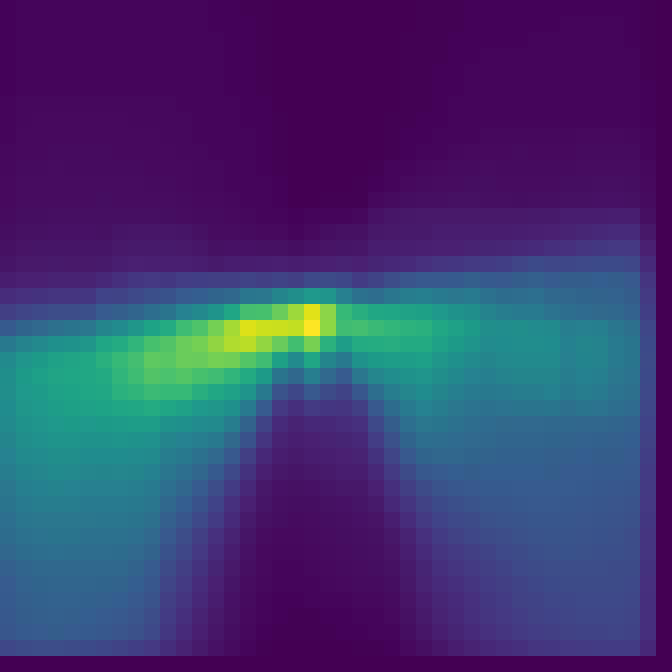}
    \caption{}
    \label{fig:region_scores_kitti}
  \end{subfigure}
  \begin{subfigure}{0.3\linewidth}
    \centering
      \includegraphics[width=0.9\linewidth]{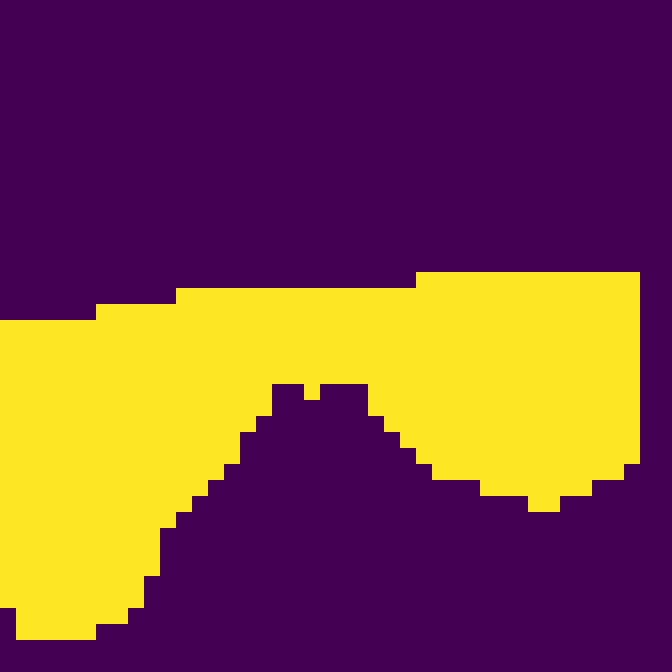}
    \caption{}
    \label{fig:static_mask_kitti_top30}
  \end{subfigure}
  \caption{An example of generating static mask from ground truth bounding boxes in KITTI train set, where 
  (a) shows the accumulated heatmap ($\mathcal{H}$), (b) shows region-wise scores, with heatmap values accumulated in regions of 16 $\times$16 pixels, and (c) shows our static mask with top 30\% regions}
  \label{fig:static_mask}
\end{figure}

\noindent\textbf{Static Mask:}
The objective of a static mask is to identify regions where objects appear more frequently within a dataset. For detection tasks, this information is readily available from the ground truth bounding boxes in the training set. We leverage this information to get an accumulated heatmap ($\mathcal{H}$) of the objects using Algorithm \ref{alg:heatmap}. As observed in the accumulated heatmap for the KITTI training set as shown in Figure \ref{fig:heatmap_kitti}, objects mostly appear along the roads for an autonomous driving task. In addition, there are some regions where objects never appear, such as in the sky. The pixel-wise values from $\mathcal{H}$ are accumulated to get region-wise scores (Figure \ref{fig:region_scores_kitti}). A region size of 16$\times$16 pixels is chosen in accordance with the patch size in the VitDet \cite{Li2022vitdet} model backbone. Based on the static keep rate ($k_{s}$) in the mask, which is a hyperparameter in our design, top-$k$ patches are selected based on region scores for further processing, where $k=int(k_{s}{*}N)$ and $N$ is the total number of input patches or tokens. 

\begin{algorithm}
\caption{Generating Accumulated Heatmap}
\label{alg:heatmap}
\begin{algorithmic}[1]
\Require{(H,W): image height and width; train\_set} 
\Ensure $\mathcal{H}$: Accumulated heatmap of objects in train set
\Statex
  \State {$\mathcal{H}$ $\gets$ {$Zeros$(H,W)}}
    \For{minibatch in (train\_set)}                    
\State{image, annotation $\gets$ minibatch}
\For{bbox in annotation['boxes']}
            \State{x1, y1, x2, y2 $\gets$ bbox}
            \State{$\mathcal{H}$(y1:y2,x1:x2) += 1}
        \EndFor
    \EndFor
\end{algorithmic}
\end{algorithm}

\vspace{3mm}
\noindent\textbf{Dynamic Mask:}
While static mask gives a general notion of regions where objects appear more frequently, for video tasks, objects are likely to almost surely appear in regions where they appeared in the previous frame. Therefore, we use the bounding boxes for detected objects in the previous frame to construct our dynamic mask for the current frame. Although objects move slightly between frames, this dislocation can often be covered by a region size of 16 pixels, as demonstrated in our experiments. 

\begin{figure}[htbp]
  \centering
  \begin{subfigure}{0.3\linewidth}
  \centering
    \includegraphics[width=0.9\linewidth]{images/static_mask_kitti_top30.png}
    \caption{}
    \label{fig:static_kitti}
  \end{subfigure}
  \begin{subfigure}{0.3\linewidth}
    \centering
      \includegraphics[width=0.9\linewidth]{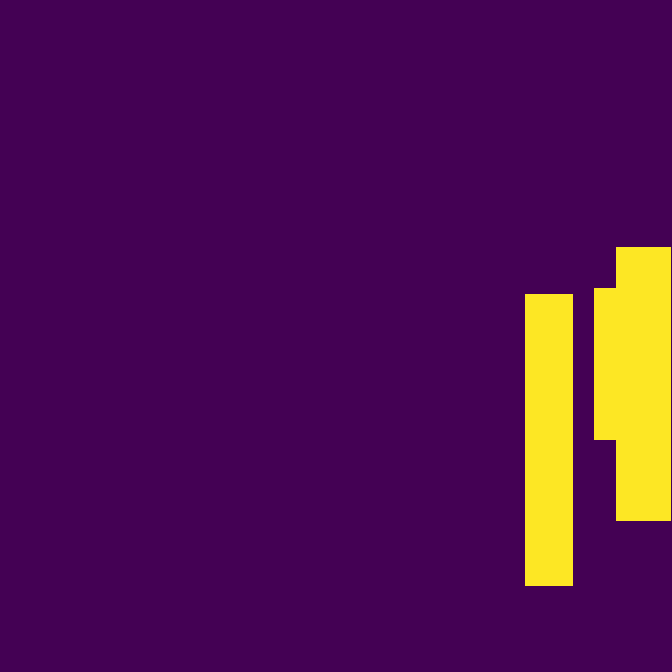}
    \caption{}
    \label{fig:dynamic_kitti}
  \end{subfigure}
  \begin{subfigure}{0.3\linewidth}
\centering
      \includegraphics[width=0.9\linewidth]{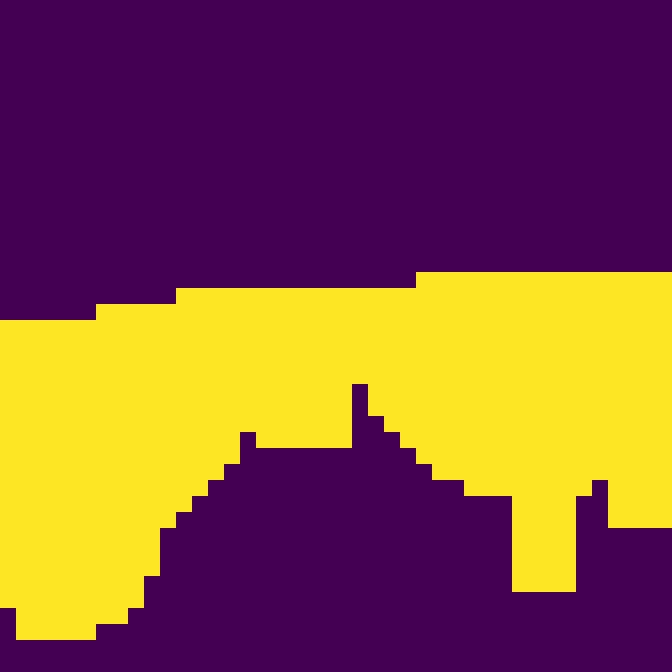}    \caption{}
    \label{fig:combined_kitti}
  \end{subfigure}
  \caption{An example of generating combined mask from static and dynamic mask in KITTI. Union on (a) static mask with top 30\% regions and (b) dynamic mask from bounding boxes of the last frame generated a (c) combined mask with top 35\% regions.}
  \label{fig:short}
\end{figure}

\begin{figure*}
    \centering
    \includegraphics[width=0.84\textwidth]{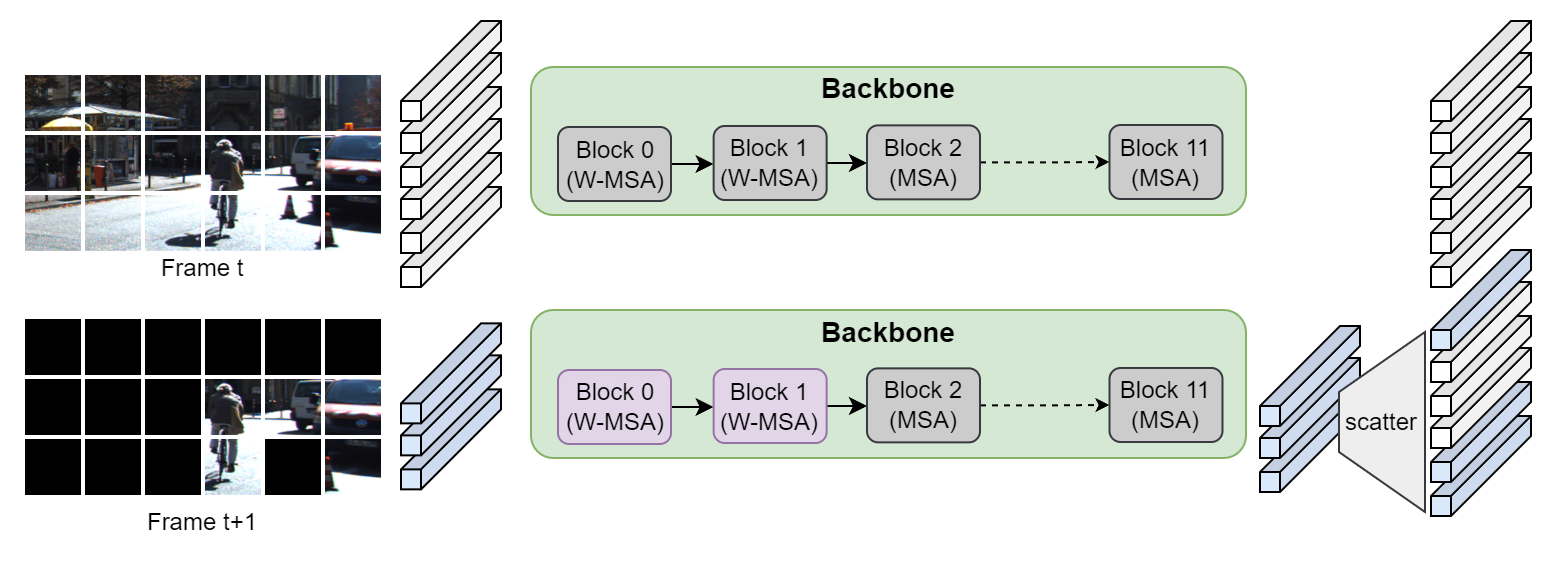}
    \caption{Inference framework of MaskVD with windowed ViT backbone. Input frame $t$ is fully processed, generating a set of reference tokens at the output. Input frame $t+1$ is masked following our proposed approach, feeding only a few tokens at the input, and the processed tokens updating the corresponding locations of the reference output. Figure \ref{fig:window_attn} shows W-MSA block blocks for masked frames.}
    \label{fig:model_arch}
\end{figure*}

\noindent\textbf{Combined Input Mask:}
We take a union of regions selected by our static and dynamic masks to obtain our combined mask. This mask is applied at the input to our detection model to all frames except the ones that are fully processed. Periodically, we process the entire frame without masking to account for new objects and to prevent the accumulation of errors in the mask from detected bounding boxes. For example, for a periodicity $P$, another hyperparameter in our design, the first frame in a window of every $P$ frames is fully processed while the subsequent $(P{-}1)$ frames are masked. The mask for frame $t$ can be completely estimated from the information available from frame $(t{-}1)$. Hence, given a custom hardware that can take advantage of the masked regions by skipping them altogether from the readout stage, our approach can significantly improve the computational efficiency. Our masks are determined only at the input to the network, unlike delta-based processing \cite{Dutson_ICCV_2023} which performs gating at each layer in the network.

\subsection{Masked Processing in Detection Backbone}
\subsubsection{Transformer Backbone}
The input mask determines the locations for the masked frames at which the tokens should be re-computed. Based on the input mask, the tokens are gathered and fed to the encoder blocks along with their token locations, as shown in Figure \ref{fig:model_arch}. The output of the last processed frame is stored as a reference output. The processed tokens for each frame are scattered into the corresponding locations of this reference output. In essence, the new value is updated for the locations associated with the processed tokens, while the reference value is retained for the remaining locations. All locations are refreshed once every $P$ frames when the entire frame is processed. It is important to note that unlike in classification tasks, where tokens are rearranged based on their significance, preserving the token locations is crucial in detection tasks \cite{liang2022evit}.

\begin{figure}
    \centering
    \includegraphics[width=0.9\linewidth]{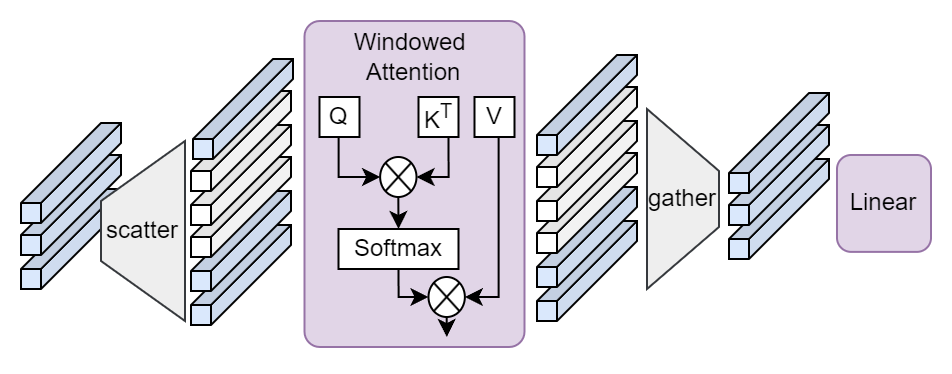}
    \caption{Windowed attention block (W-MSA) for processing masked frames. Each block maintains a set of reference tokens from the last processed frame, scattering received tokens at the input and gathering them at the output.}
    \label{fig:window_attn}
\end{figure}

When using blocks with global attention, all operations in the backbone can be seamlessly performed with dropped tokens at the input. However, windowed attention blocks encounter a dimension mismatch during window partitioning with dropped tokens. To address this, we scatter tokens to their original dimension at the input to the windowed blocks, as shown in Figure \ref{fig:window_attn}. Additionally, we maintain a reference tensor from the last processed frame at the input of each windowed-attention block, and the tokens at the input to the blocks are updated on top of this reference tensor. After attention-value computation, the tokens are gathered back according to the token locations obtained from the input mask. 

We also attempted to accelerate windowed attention by dropping windows in the $\mathbf{Q}$, 
$\mathbf{K}$, and $\mathbf{V}$ tensors. However, the latency overhead of gathering windows and scattering them post-attention computation proved to be higher than performing query-key and attention-value products with all tokens. This is possibly due the fact that highly optimized matrix multiplication operations are faster than scatter-gather operations using CUDA kernels on GPUs. 

Overall, MaskVD needs to store one tensor for each windowed attention block and the reference output, leading to a nominal memory increase over the baseline.
Furthermore, it needs no additional parameters, changes to the model architecture, or re-training and, therefore, can be easily plugged in during inference.

\noindent\textbf{Improving latency with non-windowed backbones:}
Windowed attention is primarily introduced with a goal of reducing the computational cost of global attention. However, for token-dropping methods, dimension mismatch during window partitioning creates a challenge. For non-windowed backbones, that is, ViT backbones where all blocks perform global attention, dropped tokens can be propagated through the entire backbone. They are found to be less computationally expensive than windowed backbones for the same level of sparsity (see Table \ref{tab:results}). However, windowed backbones in our case leverage features from the last computed frame, by maintaining a reference tensor for each block which boosts accuracy. Therefore, using a non-windowed backbone provides a more latency-efficient alternative at a small cost in accuracy.

\subsubsection{CNN Backbone}
Our region masking approach, masking regions of 16$\times$16 pixels, can be extended to CNN-based detection backbones with improved accuracy. Since input or activation sparsity does not provide speedup for CNNs on GPUs, we leverage the specialized kernels devised for leveraging pixel-wise sparsity in DeltaCNNs \cite{parger2022deltacnn}. DeltaCNN reduces computation by reusing previous frame outputs and performing recomputations for pixels whose delta difference from the previous frame exceeds a predetermined threshold. Pixel differences may arise even in background regions, especially for moving cameras. Region masking zeroes out delta values between frames in masked portions, significantly reducing computation and providing speedup. 

It is important to note that CNN models need fine-tuning with masked inputs to achieve improved mAP-50. However, fine-tuning is performed only once with a static mask with 50\% sparsity and this model is used
for evaluations of all combinations of static and dynamic sparsity (with no re-training).

\section{Experimental Results}
\label{sec:experiments}
\subsection{Experimental Setup}
\begin{table*}[ht!]
\small\addtolength{\tabcolsep}{-1pt}
    \centering
    \begin{tabular}{c|c|c|c|c|c|c|c|c|c}
    \toprule
    \textbf{Dataset} & \textbf{Backbone}         & \textbf{Tokens} & \textbf{Patch}         & \textbf{Period} & \textbf{Static Keep}  & \textbf{mAP-50} & \textbf{FLOPs} & \textbf{Latency} & \textbf{Memory} \\
                    &  & \textbf{Processed} & \textbf{Keep  Rate} & \textbf{($P$)} & \textbf{Rate  ($k_{s}$)} & & \textbf{(GMACs)} & \textbf{(ms)} & \textbf{(MB)} \\
    \midrule
    KITTI & Windowed & 1764 & 1.0 & - & - & 75.85 & 174.93 & 54.64 & 1173\\
    &  & 1005 & 0.57 & 8 & 0.5 & \textbf{76.04} & 110.75 & 45.43 & 1345\\
    &  & 723 & 0.41 & 8 & 0.3 & \textbf{75.94} & 85.08 & 39.55 & 1344\\
    &  & 459 & 0.26 & 8 & 0.1 & 75.21 & 63.59 & 36.45 & 1344\\
    & & 952 & 0.54 & 16 & 0.5 & \textbf{76.00} & 106.18 & 44.97 & 1344 \\
    & & 635 & 0.36 & 16 & 0.3 & \textbf{75.91}  & 79.16 & 38.14 & 1344\\
    & & 370 & 0.21 & 16 & 0.1 & \underline{74.95} & \underline{\textbf{55.67}} & \underline{\textbf{35.85}} & \underline{\textbf{1344}}\\
    \cmidrule{2-10}
    & Non- & 1764 & 1.0 & - & - & 74.81 & 208.85 & 53.88 & 1182\\
    & Windowed & 1005 & 0.57 & 8 & 0.5 & \textbf{74.18} & 105.3 & 32.42 & 1184\\
    & & 723 & 0.41 & 8 & 0.3 & 71.66 & 72.28 & 27.22 & 1185\\
    & & 952 & 0.54 & 16 & 0.5 & \underline{74.01} & \underline{98.87} & \underline{31.24} & \underline{1188}\\
    & & 635 & 0.36 & 16 & 0.3 & 71.48 & \textbf{62.59} & \textbf{25.53} & \textbf{1185}\\
    \midrule
    ImageNet-VID & Windowed & 1764 & 1.0 & - & - & 82.28 & 174.93 & 49.16 & 757\\
    & & 952 & 0.54 & 4 & 0.3 & \textbf{82.11} & 106.43 & 42.15 & 928\\
    & & 758 & 0.43 & 4 & 0.1 & 82.08 & 89.34 & 39.56 & 928\\
    & & 829 & 0.47 & 8 & 0.3 & 81.9 & 94.98 & 40.89 & 928\\
    & & 582 & 0.33 & 8 & 0.1 & \underline{81.78} & \underline{75.29} & \underline{38.95} & \underline{928}\\
    & & 388 & 0.22 & 16 & 0.0 & 81.06 & 57.56 & 37.03 & 928\\
    & & 352	& 0.2 & 24 & 0.0 & 80.51 & \textbf{54.84} & \textbf{35.71} & \textbf{928}\\
    \cmidrule{2-10}
    & Non- & 1764 & 1.0 & - & - & 79.38 & 208.85 & 54.58 & 761\\
    & Windowed & 1164 & 0.66 & 4 & 0.5 & \textbf{78.16} & 125.22 & 38.49 & 762

\\
    & & 1076 & 0.61 & 8 & 0.5 & \underline{77.77} & \underline{\textbf{114.08}} & \underline{\textbf{34.47}} & \underline{\textbf{762}}\\
    
    \bottomrule
    \end{tabular}
    \caption{Evaluation results of MaskVD using ViTDeT on ImageNet-VID \cite{russakovsky2015ImageNet} and KITTI. mAP-50 and hardware metrics like memory, latency and FLOPs are evaluated at different patch keep rates, set by varying $k_{s}$ and $P$. While FLOPs is calculated only for the backbone, latency and memory values are measured for the entire detection pipeline. The best values are highlighted in bold, while the ones that provide a good trade-off are underlined.}
    \label{tab:results}
\end{table*}

\noindent\textbf{Model:}
We evaluate our approach on the ViTDet \cite{Li2022vitdet} model, which uses a ViT-B backbone consisting of 12 blocks. We evaluate with two different configurations of the backbone: \textit{windowed}, which uses global attention in blocks 3, 6, 9 and 12 and windowed attention in the remaining 8 blocks, and \textit{non-windowed}, which uses global attention in all blocks. All video frames are rescaled and padded to a uniform size of 672{$\times$}672. The patch-embedding layer divides the input images into 16{$\times$}16 non-overlapping image patches or tokens, each of which is mapped to a token embedding vector of length 768. Windowed self-attention uses a window size of 14$\times$14 tokens.
For evaluating masking on CNN backbones, we use a SOTA object detection model, DETR \cite{carion2020detr}, which uses a ResNet-50 \cite{he2016deep} backbone. 

\vspace{3mm}
\noindent\textbf{Dataset:}
We evaluate our approach on two different datasets: ImageNet VID \cite{russakovsky2015ImageNet} and KITTI Tracking \cite{Geiger2012CVPR}. ImageNet-VID consists of 30 categories of objects, carefully selected from the 200 categories in ImageNet Detection task. The results are reported on the validation set, which consists of 639 sequences, each containing up to 2895 frames. KITTI \cite{Geiger2012CVPR} is an autonomous driving dataset captured by a camera mounted on a car moving through traffic. The videos are recorded at 10 frames per second (FPS) consisting of large inter-frame motions. It contains 21 training sequences and 29 test sequences, for which annotations are not available. The training sequence is split into halves for training and validation, and evaluation is performed on the validation set on 3 classes: car, pedestrian, and cyclists \cite{zhou2020tracking}. 

\vspace{3mm}
\noindent\textbf{Evaluation Metrics:}
We evaluate object detection performance using mean average precision with an IoU threshold of 0.5, which is indicated as mAP-50. FLOPs are measured for the detection backbones. The latency and memory values are measured for the entire detection pipeline on NVIDIA RTX A6000 GPUs.

\vspace{3mm}
\noindent\textbf{Training Hyperparameters:}
We use the pre-trained ViTDet \cite{Li2022vitdet} model with windowed backbone from \cite{Dutson_ICCV_2023} for the ImageNet-VID dataset. For KITTI, ViTDet \cite{Li2022vitdet} trained on ImageNet-VID is fine-tuned for 8 epochs using the AdamW optimizer on the KITTI training set, with an initial learning rate of $10^{-5}$, 3 warmup epochs, and a cosine learning rate decay.  
ViTDeT models with non-windowed ViT-B backbones start with ImageNet pre-trained ViT-B models and are fine-tuned for 5 epochs on ImageNet-VID and 8 epochs on KITTI following similar configurations as stated above. DETR \cite{carion2020detr} evaluation is performed on a model fine-tuned on KITTI with a static mask with 50\% sparsity. Fine-tuning is performed for 50 epochs following the training parameters specified in \cite{carion2020detr} with an initial learning rate of $10^{-5}$. 
All evaluation results in this paper are reported using MaskVD during inference without re-training. 

\begin{figure*}
    \centering
    \includegraphics[width=\linewidth]{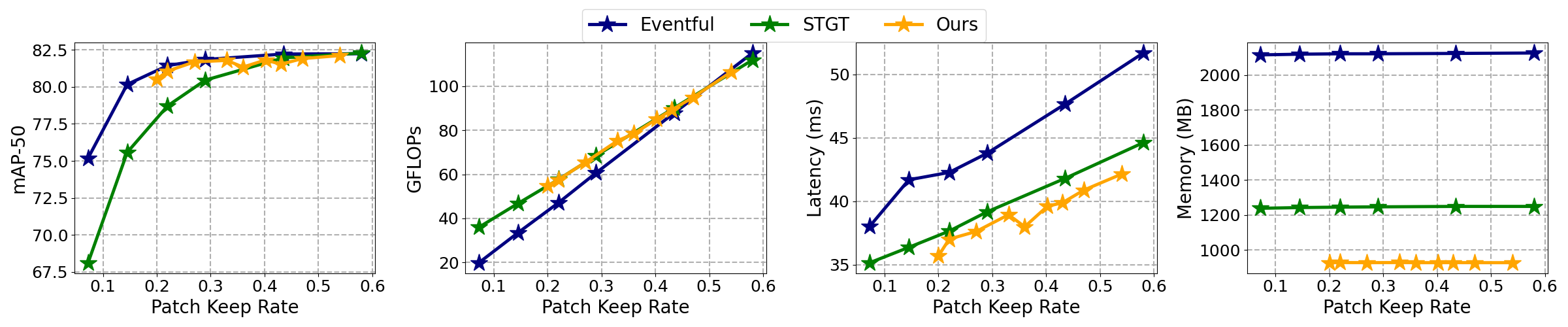}
    \caption{Comparison of our approach with Eventful-Transformers \cite{Dutson_ICCV_2023} and STGT \cite{liSpatiotemporalGated2021} on ImageNet-VID in terms of mAP-50, FLOPs, latency, and memory. STGT shows a drop in mAP at low patch keep rates, whereas Eventful-Transformer significantly increases the memory overhead. Our approach significantly reduces latency and memory over existing methods while maintaining mAP-50.}
    \label{fig:eventful}
\end{figure*}

\subsection{Results and Analysis}
Table \ref{tab:results} presents the evaluation results for our approach. We set the patch keep rate, that is, the fraction of patches processed by varying the hyperparameters, $P$, the period after which a frame is completely processed, and $k_{s}$, the keep rate of the static mask. 

We observe that an input region sparsity of up to 64\% on the KITTI dataset, in fact, improves mAP-50 over the baseline while reducing FLOPs by 2.2$\times$ and latency by 1.4$\times$. We are able to achieve FLOPs and latency reduction of 3.14$\times$ and 1.5$\times$, respectively, on KITTI with mAP-50 degradation of only 0.9\%. The non-windowed ViT-B backbone shows further improvement in FLOPs and latency, although FLOPs in the baseline non-windowed model are higher than in the windowed model. Our approach achieves 4.8$\times$ FLOPs and 1.7$\times$ latency reduction with 0.8\% mAP-50 reduction over the baseline with a non-windowed backbone. 

For ImageNet-VID dataset, we improve FLOPs and latency by 2.3$\times$ and 1.3$\times$ with mAP-50 degradation of only 0.5\% (underlined in Table). Our approach can reduce FLOPs up to 3.2$\times$ and latency up to 1.4$\times$ with $<$2\% degradation in mAP. Notably, a lower $P$ achieves better mAP even with a lower keep rate. Non-windowed ViT-B backbone shows higher reduction, improving latency by 1.6$\times$ with only 39\% input sparsity. However, the non-windowed backbone suffers a slightly higher degradation in mAP. We observe a small increase in memory of $\sim$170 MB over baseline for windowed backbones, whereas for non-windowed backbones, there is no additional memory requirement.

\vspace{3mm}
\noindent\textbf{Do we need both static and dynamic masks?} In Table \ref{tab:mask}, we compare results using only static, only dynamic, and combined masks for similar patch keep rates. Static mask alone suffers a considerable drop in mAP. Although dynamic mask alone performs reasonably well for ImageNet-VID, as shown in Table \ref{tab:results} for static keep rate 0.0 (indicating the static mask is not used), it shows a drop in mAP for KITTI, possibly due to higher inter-frame motion. Combining static and dynamic masks provides a tunable patch keep rate and an effective method for patch selection, achieving close to the baseline performance of 75.85\% at $\sim$80\% sparsity.

\begin{table}[htbp]
\small\addtolength{\tabcolsep}{-1pt}
    \centering
    \begin{tabular}{c|c|c}
    \toprule
    \textbf{Patch Keep Rate} & \textbf{Masking} & \textbf{mAP-50} \\
    \midrule
     0.21 & Static & 63.85 \\
     0.20 & Dynamic & 73.12  \\
     0.21 & Combined & \textbf{74.95} \\
    \bottomrule
    \end{tabular}
    \caption{Comparison between static, dynamic and combined masks ($P$=16, $k_{s}$=0.1) on KITTI at similar patch keep rates.}
    \label{tab:mask}
\end{table}
\subsection{Comparison with SOTA}
\noindent\textbf{Comparison with video inference acceleration methods:}
In Figure \ref{fig:eventful}, we compare our approach with the SOTA methods focusing on exploiting temporal redundancy between frames for video inference acceleration, namely Eventful-Transformers~\cite{Dutson_ICCV_2023} and STGT~\cite{liSpatiotemporalGated2021}, in terms of mAP-50, latency, memory, and FLOPs. We use the STGT implementation provided in \cite{Dutson_ICCV_2023}. As shown in Figure \ref{fig:eventful}, STGT \cite{liSpatiotemporalGated2021} suffers a performance drop at high token drop rates. Our approach improves mAP-50 by $\sim$2.4\% over STGT while reducing memory by 1.4$\times$. Eventful-Transformer \cite{Dutson_ICCV_2023} can drop 85\% of input patches with a mAP degradation of only 2.12\%. Despite this leading to a significant reduction in calculated FLOPs, it does not yield a substantial latency reduction on GPUs with existing kernels.
Our approach achieves mAP-50 on par with Eventful-Transformer while providing a reduction of 2.3$\times$ in memory and $\sim$1.14$\times$ in latency with only a slight increase in FLOPs.

To appreciate the significant drop in memory provided by our approach, we quantify the memory associated with token gating in existing methods. Token gates and buffers are present at the input and output of each layer within a transformer block, as discussed in Section \ref{sec:related_work}. For an input image size 672$\times$672, each token gate and buffer stores a set of 1764 768-dimensional tokens that takes $\sim$5.4 MB. Each transformer block in \cite{Dutson_ICCV_2023, liSpatiotemporalGated2021} contains a total of 8 token gates and buffer modules, storing ${\sim}$43.2 MB for each block. Additionally, \cite{Dutson_ICCV_2023} requires storing two product tensors of dimension 1764{$\times$}1764{$\times$}12 and 1764{$\times$}768 for accelerating the query-key and attention-value product, adding another 155 MB, increasing memory requirement by 198 MB per block and 2376 MB for a total of 12 blocks. The memory issue is further aggravated with an increase in input dimension. In contrast, our approach needs to store a single reference tensor containing 1764 768-dimensional tokens for input size 672$\times$672 for the windowed transformer blocks, as shown in Figure \ref{fig:window_attn}, adding 5.4 MB per block and a total of 43.2 MB for the 8 windowed blocks, over a 55$\times$ decrease in memory overhead. 

To appreciate the latency benefit provided by our approach, note that the additional operations due to the token gate and buffers also reduce the latency benefits of token reduction \cite{Dutson_ICCV_2023}. Our approach not only drastically reduces the memory requirement but also limits scatter-gather operations to only once per windowed block, thereby improving the system latency. 

\begin{figure}
    \centering
    \includegraphics[width=0.7\linewidth]{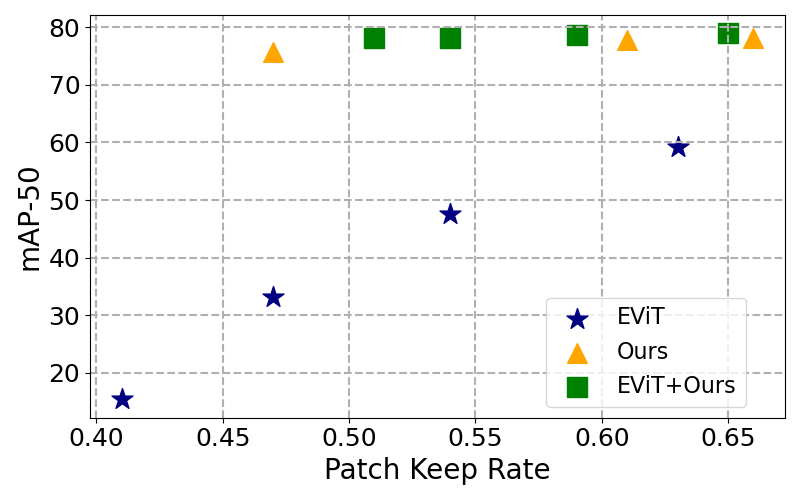}
    \caption{Comparison of our approach and EViT on ImageNet-VID Dataset on ViTDET with non-windowed Vit-B backbone}
    \vspace{-2mm}
    \label{fig:evit}
\end{figure}

\vspace{4mm}
\noindent\textbf{Comparison with token-dropping methods:}
In this section, we evaluate how token-dropping techniques like EViT \cite{liang2022evit}, designed for classification tasks, scale to the domain of object detection via interpolation with zeros. We use a non-windowed ViT-B backbone similar to \cite{liang2022evit, bolya2022tokenmerge}. Our results, shown in Figure \ref{fig:evit}, indicate that our method yields significantly higher mAP-50 values than EViT \cite{liang2022evit} and underscores the benefits of using extracted features from the previous frame. It is also worth noting that the attention-based token-dropping approach used in EViT is independent of our input masking method and can be effectively combined with our approach to achieve slightly better performance. In particular, when combining EViT with our approach for the same level of sparsity, we observe an improvement of ${\sim}0.8\%$ in mAP (as shown in Figure~\ref{fig:evit}).

\subsection{Evaluation on CNN Backbones}
\begin{figure}
    \centering
    \includegraphics[width=0.7\linewidth]{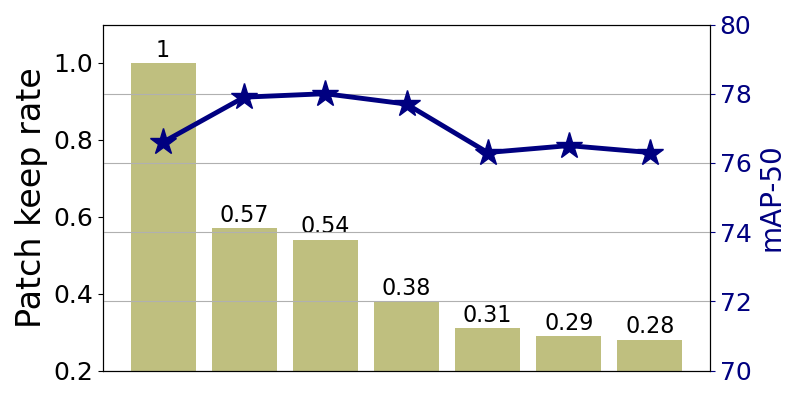}
    \caption{mAP-50 of DETR with a ResNet-50 backbone on KITTI with different input mask sparsity}
    \label{fig:detr_map}
\end{figure}

We showcase the results of our proposed region masking approach on DETR \cite{carion2020detr}, which uses a ResNet-50 \cite{he2016deep} backbone in Figure \ref{fig:detr_map}. The evaluation is performed on a DETR model trained with a static input mask with 50\% sparsity. The input mask sparsity is tuned by varying the period ($P$) in the range from 8 to 32 and ($k_{s}$) from 0.5 to 0.2. We observe that an input patch keep rate of 55\% improves mAP-50 by 1.4\% over baseline. In fact, mAP improves over baseline up to a keep rate of $\sim$40\%. We observe a drop of ${<}$0.3\% in mAP for lower patch keep rates.

\begin{figure}
    \centering
    \includegraphics[width=0.7\linewidth]{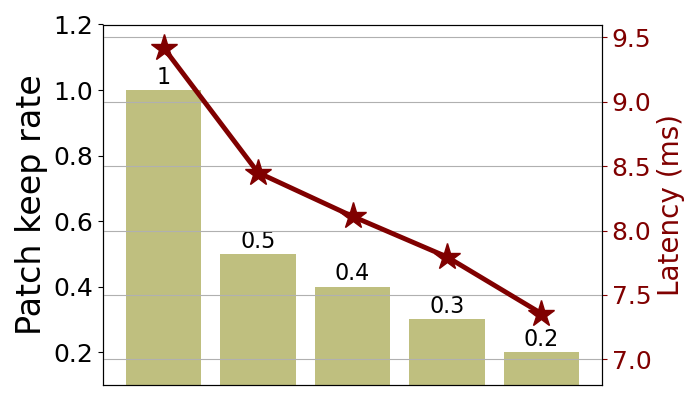}
    \caption{Latency for DeltaCNN \cite{parger2022deltacnn} based ResNet backbone with different input mask sparsity}
    \label{fig:detr_latency}
\end{figure}

As discussed earlier, activation sparsity does not directly translate to latency benefits on GPUs. DeltaCNN \cite{parger2022deltacnn} designs specialized CUDA kernels to obtain actual speedup on GPUs using pixel-level sparsity. By leveraging these kernels, we demonstrate that masking the input can further reduce computations and yield significant speedup over regular DeltaCNN. As shown in Figure \ref{fig:detr_latency}, an input mask with a keep rate 0.2 can reduce latency by 1.3$\times$ over DeltaCNN with a threshold of 0.5.

\section{Conclusions}
\label{sec:conclusion}
Video images undergo very little change over large portions of the frame leading to redundant computations for frame-based video tasks. We propose MaskVD, a region masking strategy for efficient video object detection that can be directly plugged in during inference without re-training. Our region masking approach considers the semantic information in images and the temporal redundancy across frames, and can be applied to both CNNs and ViTs alike. Unlike prior works, region selection using our strategy is sufficient to be performed only at the input of the detection pipeline and does not require re-selection in the intermediate layers. We demonstrate promising results with improved mAP for both ViT and CNN backbones. For ViT backbones, in contrast to existing approaches, our method stores a single tensor for every windowed transformer block to reuse extracted features from previous frames, ensuring improved mAP-50 with minimum memory overhead. Our method significantly improves memory and latency over existing approaches, which, although showing significant reduction in FLOPs, yield limited latency benefits and high memory overhead. For CNN backbones, we leverage specialized CUDA kernels capable of exploiting pixel-wise sparsity. Region masking zeroes out delta differences in large portions of the frame, reducing computations and providing significant speedup.
{\small
\bibliographystyle{ieee_fullname}
\bibliography{egbib}

\begin{thebibliography}{10}\itemsep=-1pt

\bibitem{bolya2022tokenmerge}
Daniel Bolya, Cheng-Yang Fu, Xiaoliang Dai, Peizhao Zhang, Christoph Feichtenhofer, and Judy Hoffman.
\newblock Token merging: Your {ViT} but faster.
\newblock {\em arXiv preprint arXiv:2210.09461}, 2022.

\bibitem{nuscenes2019}
Holger Caesar, Varun Bankiti, Alex~H. Lang, Sourabh Vora, Venice~Erin Liong, Qiang Xu, Anush Krishnan, Yu Pan, Giancarlo Baldan, and Oscar Beijbom.
\newblock nuscenes: A multimodal dataset for autonomous driving.
\newblock {\em arXiv preprint arXiv:1903.11027}, 2019.

\bibitem{carion2020detr}
Nicolas Carion, Francisco Massa, Gabriel Synnaeve, Nicolas Usunier, Alexander Kirillov, and Sergey Zagoruyko.
\newblock End-to-end object detection with transformers.
\newblock In {\em European conference on computer vision}, pages 213--229. Springer, 2020.

\bibitem{cavigelliCBinferChangebased2017}
Lukas Cavigelli, Philippe Degen, and Luca Benini.
\newblock {CBinfer}: Change-based inference for convolutional neural networks on video data.
\newblock In {\em Proceedings of the 11th International Conference on Distributed Smart Cameras}. Association for Computing Machinery, September 2017.

\bibitem{chen2017deeplab}
Liang-Chieh Chen, George Papandreou, Iasonas Kokkinos, Kevin Murphy, and Alan~L Yuille.
\newblock {DeepLab}: Semantic image segmentation with deep convolutional nets, atrous convolution, and fully connected crfs.
\newblock {\em IEEE transactions on pattern analysis and machine intelligence}, 40(4):834--848, 2017.

\bibitem{chen2023sparsevit}
Xuanyao Chen, Zhijian Liu, Haotian Tang, Li Yi, Hang Zhao, and Song Han.
\newblock {SparseViT}: Revisiting activation sparsity for efficient high-resolution vision transformer.
\newblock In {\em IEEE/CVF Conference on Computer Vision and Pattern Recognition (CVPR)}, 2023.

\bibitem{chen2022mobile}
Yinpeng Chen, Xiyang Dai, Dongdong Chen, Mengchen Liu, Xiaoyi Dong, Lu Yuan, and Zicheng Liu.
\newblock Mobile-former: Bridging {MobileNet} and transformer.
\newblock In {\em Proceedings of the IEEE/CVF Conference on Computer Vision and Pattern Recognition}, pages 5270--5279, 2022.

\bibitem{choromanski2020rethinking}
Krzysztof Choromanski, Valerii Likhosherstov, David Dohan, Xingyou Song, Andreea Gane, Tamas Sarlos, Peter Hawkins, Jared Davis, Afroz Mohiuddin, Lukasz Kaiser, et~al.
\newblock Rethinking attention with performers.
\newblock {\em arXiv preprint arXiv:2009.14794}, 2020.

\bibitem{in-sensor1}
G. Datta et~al.
\newblock A processing-in-pixel-in-memory paradigm for resource-constrained tinyml applications.
\newblock {\em Scientific Reports}, 12, 2022.

\bibitem{in-sensor2}
Gourav Datta, Zeyu Liu, Md~Abdullah-Al Kaiser, Souvik Kundu, Joe Mathai, Zihan Yin, Ajey~P. Jacob, Akhilesh~R. Jaiswal, and Peter~A. Beerel.
\newblock In-sensor \& neuromorphic computing are all you need for energy efficient computer vision.
\newblock In {\em ICASSP 2023 - 2023 IEEE International Conference on Acoustics, Speech and Signal Processing (ICASSP)}, pages 1--5, 2023.

\bibitem{dosovitskiy2020image}
Alexey Dosovitskiy, Lucas Beyer, Alexander Kolesnikov, Dirk Weissenborn, Xiaohua Zhai, Thomas Unterthiner, Mostafa Dehghani, Matthias Minderer, Georg Heigold, Sylvain Gelly, Jakob Uszkoreit, and Neil Houlsby.
\newblock An image is worth 16x16 words: Transformers for image recognition at scale.
\newblock In {\em International Conference on Learning Representations}, 2021.

\bibitem{optical_flow_motion}
Alexey Dosovitskiy, Philipp Fischer, Eddy Ilg, Philip Häusser, Caner Hazirbas, Vladimir Golkov, Patrick van~der Smagt, Daniel Cremers, and Thomas Brox.
\newblock Flownet: Learning optical flow with convolutional networks.
\newblock In {\em 2015 IEEE International Conference on Computer Vision (ICCV)}, pages 2758--2766, 2015.

\bibitem{Dutson_ICCV_2023}
Matthew Dutson, Yin Li, and Mohit Gupta.
\newblock Eventful {Transformers}, leveraging temporal redundancy in vision transformers.
\newblock In {\em Proceedings of the IEEE International Conference on Computer Vision (ICCV)}, 2023.

\bibitem{fayyaz2022ats}
Mohsen Fayyaz, Soroush Abbasi~Kouhpayegani, Farnoush Rezaei~Jafari, Eric Sommerlade, Hamid~Reza Vaezi~Joze, Hamed Pirsiavash, and Juergen Gall.
\newblock Adaptive token sampling for efficient vision transformers.
\newblock {\em European Conference on Computer Vision (ECCV)}, 2022.

\bibitem{Geiger2012CVPR}
Andreas Geiger, Philip Lenz, and Raquel Urtasun.
\newblock Are we ready for autonomous driving? the {KITTI} vision benchmark suite.
\newblock In {\em Conference on Computer Vision and Pattern Recognition (CVPR)}, 2012.

\bibitem{habibianSkipconvolutionsEfficient2021}
Amirhossein Habibian, Davide Abati, Taco~S. Cohen, and Babak~Ehteshami Bejnordi.
\newblock Skip-convolutions for efficient video processing.
\newblock In {\em Proceedings of the IEEE/CVF Conference on Computer Vision and Pattern Recognition (CVPR)}, pages 2695--2704, June 2021.

\bibitem{he2016deep}
Kaiming He, Xiangyu Zhang, Shaoqing Ren, and Jian Sun.
\newblock Deep residual learning for image recognition.
\newblock In {\em Proceedings of the IEEE conference on computer vision and pattern recognition}, pages 770--778, 2016.

\bibitem{cifar}
Alex Krizhevsky and Geoffrey Hinton.
\newblock Learning multiple layers of features from tiny images.
\newblock Technical report, Citeseer, 2009.

\bibitem{liSpatiotemporalGated2021}
Yawei Li, Babak~Ehteshami Bejnordi, Bert Moons, Tijmen Blankevoort, Amirhossein Habibian, Radu Timofte, and Luc Van~Gool.
\newblock Spatio-temporal gated transformers for efficient video processing.
\newblock In {\em Advances in Neural Information Processing Systems Workshops}, 2021.

\bibitem{Li2022vitdet}
Yanghao Li, Hanzi Mao, Ross~B. Girshick, and Kaiming He.
\newblock Exploring plain vision transformer backbones for object detection.
\newblock {\em ArXiv}, abs/2203.16527, 2022.

\bibitem{li2022efficientformer}
Yanyu Li, Geng Yuan, Yang Wen, Ju Hu, Georgios Evangelidis, Sergey Tulyakov, Yanzhi Wang, and Jian Ren.
\newblock {EfficientFormer}: Vision transformers at {MobileNet} speed.
\newblock {\em Advances in Neural Information Processing Systems}, 35:12934--12949, 2022.

\bibitem{liang2022evit}
Youwei Liang, Chongjian Ge, Zhan Tong, Yibing Song, Jue Wang, and Pengtao Xie.
\newblock Not all patches are what you need: Expediting vision transformers via token reorganizations.
\newblock {\em arXiv preprint arXiv:2202.07800}, 2022.

\bibitem{lin2014microsoftcoco}
Tsung-Yi Lin, Michael Maire, Serge Belongie, James Hays, Pietro Perona, Deva Ramanan, Piotr Doll{\'a}r, and C~Lawrence Zitnick.
\newblock Microsoft {COCO}: Common objects in context.
\newblock In {\em Computer Vision--ECCV 2014: 13th European Conference, Zurich, Switzerland, September 6-12, 2014, Proceedings, Part V 13}, pages 740--755. Springer, 2014.

\bibitem{parger2023motiondeltacnn}
Mathias Parger, Chengcheng Tang, Thomas Neff, Christopher~D Twigg, Cem Keskin, Robert Wang, and Markus Steinberger.
\newblock {MotionDeltaCNN}: Sparse {CNN} inference of frame differences in moving camera videos with spherical buffers and padded convolutions.
\newblock In {\em Proceedings of the IEEE/CVF International Conference on Computer Vision}, pages 17292--17301, 2023.

\bibitem{parger2022deltacnn}
Mathias Parger, Chengcheng Tang, Christopher~D Twigg, Cem Keskin, Robert Wang, and Markus Steinberger.
\newblock {DeltaCNN}: End-to-end {CNN} inference of sparse frame differences in videos.
\newblock {\em CVPR 2022}, June 2022.

\bibitem{pinkhan2021jetcas}
Reid Pinkham et~al.
\newblock Near-sensor distributed {DNN} processing for augmented and virtual reality.
\newblock {\em IEEE JETCAS}, 11(4), 2021.

\bibitem{Ronneberger2015UNetCN}
Olaf Ronneberger, Philipp Fischer, and Thomas Brox.
\newblock U-net: Convolutional networks for biomedical image segmentation.
\newblock In Nassir Navab, Joachim Hornegger, William~M. Wells, and Alejandro~F. Frangi, editors, {\em Medical Image Computing and Computer-Assisted Intervention -- MICCAI 2015}, pages 234--241, Cham, 2015. Springer International Publishing.

\bibitem{russakovsky2015ImageNet}
Olga Russakovsky, Jia Deng, Hao Su, Jonathan Krause, Sanjeev Satheesh, Sean Ma, Zhiheng Huang, Andrej Karpathy, Aditya Khosla, Michael Bernstein, Alexander~C. Berg, and Li Fei-Fei.
\newblock {ImageNet} large scale visual recognition challenge.
\newblock {\em International Journal of Computer Vision}, 115(3):211--252, April 2015.

\bibitem{sarkarECV24}
Sreetama Sarkar, Souvik Kundu, Kai Zheng, and Peter Beerel.
\newblock Block selective reprogramming for on-device training of vision transformers.
\newblock In {\em Proceedings of the IEEE/CVF Conference on Computer Vision and Pattern Recognition (CVPR) Workshops}, 2024.

\bibitem{optical_flow_network}
Karen Simonyan and Andrew Zisserman.
\newblock Two-stream convolutional networks for action recognition in videos.
\newblock In Z. Ghahramani, M. Welling, C. Cortes, N. Lawrence, and K.Q. Weinberger, editors, {\em Advances in Neural Information Processing Systems}, volume~27. Curran Associates, Inc., 2014.

\bibitem{wangEfficientVideo2022}
Junke Wang, Xitong Yang, Hengduo Li, Li Liu, Zuxuan Wu, and Yu-Gang Jiang.
\newblock Efficient video transformers with spatial-temporal token selection.
\newblock In Shai Avidan, Gabriel Brostow, Moustapha Ciss{\'e}, Giovanni~Maria Farinella, and Tal Hassner, editors, {\em Proceedings of the European Conference on Computer Vision (ECCV)}, pages 69--86, Cham, 2022. Springer Nature Switzerland.

\bibitem{wang2020linformer}
Sinong Wang, Belinda~Z Li, Madian Khabsa, Han Fang, and Hao Ma.
\newblock Linformer: Self-attention with linear complexity.
\newblock {\em arXiv preprint arXiv:2006.04768}, 2020.

\bibitem{compressed_video}
Chao-Yuan Wu, Manzil Zaheer, Hexiang Hu, R. Manmatha, Alexander~J. Smola, and Philipp Krähenbühl.
\newblock Compressed video action recognition.
\newblock In {\em 2018 IEEE/CVF Conference on Computer Vision and Pattern Recognition}, pages 6026--6035, 2018.

\bibitem{xu2018deepcache}
Mengwei Xu, Mengze Zhu, Yunxin Liu, Felix~Xiaozhu Lin, and Xuanzhe Liu.
\newblock {DeepCache}: Principled cache for mobile deep vision.
\newblock In {\em Proceedings of the 24th annual international conference on mobile computing and networking}, pages 129--144, 2018.

\bibitem{zhang2023sal}
Yuke Zhang, Dake Chen, Souvik Kundu, Chenghao Li, and Peter~A Beerel.
\newblock {SAL-ViT}: Towards latency efficient private inference on {ViT} using selective attention search with a learnable softmax approximation.
\newblock In {\em Proceedings of the IEEE/CVF International Conference on Computer Vision}, pages 5116--5125, 2023.

\bibitem{zhou2020tracking}
Xingyi Zhou, Vladlen Koltun, and Philipp Kr{\"a}henb{\"u}hl.
\newblock Tracking objects as points.
\newblock {\em ECCV}, 2020.

\end{thebibliography}
}

\end{document}